\documentclass[10pt,twocolumn,letterpaper]{article}

 \usepackage[pagenumbers]{cvpr} 

\usepackage{microtype}

 \renewcommand{\paragraph}[1]{\vspace{.5em}\noindent\textbf{#1.}}

 \usepackage{xspace}

 \usepackage{soul}
 \setuldepth{foobar}

\usepackage{makecell}
\usepackage{booktabs}

\definecolor{cvprblue}{rgb}{0.21,0.49,0.74}
\usepackage[pagebackref,breaklinks,colorlinks,allcolors=cvprblue]{hyperref}

\def\confName{CVPR}
\def\confYear{2027}

\title{PAIR: Depiction-Invariant Representation Learning for Optical Chemical Structure Recognition}

\author{
Christina Yi Jin\textsuperscript{1}\quad
Yuan Jin\textsuperscript{1}\quad
Ziyang Chen\textsuperscript{2}\\
Baochen Wang\textsuperscript{2}\quad
Qing Zhang\textsuperscript{1,*}\\[0.5em]
\textsuperscript{1}Research Center for Frontier Fundamental Studies\\
\textsuperscript{2}Research Center for Scientific Data Hub\\
Zhejiang Lab\\
{\tt\small
\{cyj,jin.yuan,chenzy,wangbc,qing.zhang\}@zhejianglab.org}\\[0.3em]
{\small \textsuperscript{*}Corresponding author:
qing.zhang@zhejianglab.org}
}

\begin{document}
\maketitle
\begin{abstract}
Optical chemical structure recognition (OCSR) must recover the
same molecular structure from drawings that vary widely in style,
layout, and image quality. Several leading methods address this
task through explicit atom and bond prediction, supervising the
structure recovered from each image. We explore a complementary
form of structural supervision: encouraging representations of
different depictions of the same molecule to agree, so that
recognition depends less on appearance and more on shared
molecular identity. We propose Paired-view Alignment for Invariant Recognition (PAIR), a framework that learns representations insensitive to variations in molecular depiction while preserving distinctions between chemical structures. PAIR achieves this through shared learnable queries and cross-view contrastive learning on paired real and rendered molecular images. To complement the query-based representations, residual fusion preserves fine-grained spatial evidence for accurate SMILES decoding. Evaluated on six OCSR
benchmarks with three diverse backbones, PAIR consistently
improves recognition performance, with PAIR-2B achieving a mean
exact molecular accuracy of 85.92\%. Specifically, on the task of
Markush structure recognition, PAIR-2B surpasses the current
state-of-the-art OCSR model by 10 percentage points.
Further analyses demonstrate improved cross-view alignment
and retrieval, while ablation studies support the contributions
of the proposed components. 
These results confirm that learning depiction invariance provides effective structural supervision that complements molecular reconstruction for robust OCSR.
\end{abstract}    

\section{Introduction}
\label{sec:intro}

Optical chemical structure recognition (OCSR) converts molecular
drawings into machine-readable structures, enabling computational
search and analysis of chemical information in scientific documents.
Although the goal is to recover the underlying molecular graph,
visual depictions vary substantially in layout, typography, and
image quality. Learning molecular identity across these variations
is therefore essential for reliable OCSR.

Existing OCSR methods explicitly model molecular structure.
For example, MolScribe predicts atoms and bonds~\cite{qian2023molscribe}, while MolGrapher employs graph-based
recognition and synthetic training data \cite{morin2023molgrapher}.
These approaches encourage reasoning about molecular components
and connectivity rather than relying entirely on implicit
image-to-SMILES translation. However, explicit structural
supervision does not necessarily constrain the encoder's
dependence on drawing style. Models may recognize familiar
depictions while remaining sensitive to unfamiliar fonts,
layouts, or image degradation.

We focus on learning representations organized around molecular
identity across visual appearances.
Different depictions of the same molecule provide complementary
observations of a shared chemical graph. Their shared identity
provides implicit structural supervision: representations should
remain consistent when appearance changes without altering the
graph, yet distinguish chemically different structures.
Aligning these observations while retaining the information
needed for SMILES reconstruction therefore encourages embeddings
that reflect the underlying molecular structure.
This motivates the following idealized proposition:
\begin{quote}
A representation uniquely identifies a molecular graph if it
is invariant across its depictions and discriminative across
distinct graphs.
\end{quote}

We propose Paired-view Alignment for Invariant Recognition
(PAIR) for OCSR. Paired real and rendered molecular images
are processed by a shared image encoder and learnable-query
module to produce compact visual token sequences.
A query-wise contrastive objective encourages alignment
across depiction variations while separating different
molecular identities. A shared autoregressive decoder
predicts the same SMILES target from either view,
encouraging preservation of chemical information.
At inference, only a single input image is required.

PAIR combines cross-view contrastive learning with
query-based residual refinement. Contrastive learning
promotes depiction consistency and molecular discrimination,
while residual fusion integrates query-derived representations
with spatial image features needed for accurate structural
reconstruction. Together, these components aim to balance
depiction invariance with chemical information preservation.

The main contributions are summarized as follows:
\begin{itemize}
    \item We propose \textbf{PAIR}, a paired-view
    representation learning framework that encourages
    depiction invariance while preserving chemically
    discriminative information for OCSR.

    \item We introduce \textbf{residual query refinement}, which augments
    spatial image features through a residual connection with
    query-derived representations. Cross-view contrastive learning
    encourages these representations to capture molecular identity
    across depictions, while the spatial features support chemically
    precise decoding.

    \item We validate PAIR across \textbf{six OCSR benchmarks and three diverse backbones}, consistently improving recognition performance. PAIR-2B achieves a mean exact molecular accuracy of 85.92\%
    and outperforms Mobile-V2 by 10 percentage points on Markush recognition. Representation analyses and ablations further establish the contributions of individual components and highlight the trade-off between depiction invariance and chemical fidelity.
\end{itemize}

\section{Related Work}
\label{sec:related_work}

\subsection{Optical Chemical Structure Recognition}

Existing OCSR approaches broadly fall into two categories:
image-to-sequence methods, which directly generate SMILES or
related representations, including DECIMER
\cite{rajan2021decimer,rajan2023decimer},
SwinOCSR \cite{xu2022swinocsr}, and MolParser
\cite{fang2025molparser}; and graph-based methods,
which explicitly reconstruct atoms and bonds,
including MolGrapher \cite{morin2023molgrapher}
and MolScribe \cite{qian2023molscribe}.

\paragraph{MolParser}
MolParser combines a streamlined recognition architecture
with E-SMILES and extensive synthetic and manually
annotated data. It addresses challenging real-document
depictions and Markush structures beyond conventional
SMILES representations. On WildMol-10K, it reports
76.9\% accuracy, compared with 66.4\% for MolScribe
and 45.5\% for MolGrapher, while emphasizing
inference efficiency and Markush coverage.

\paragraph{MolGrapher}
MolGrapher reconstructs molecular graphs from localized
atom and bond evidence, offering advantages for complex
structures where global sequence prediction may omit
components. However, its multi-stage pipeline depends
on accurate keypoint detection and bond candidate
generation. Errors in these stages, as well as OCR
and abbreviation expansion, may propagate through
subsequent reconstruction.

\paragraph{MolScribe}
MolScribe combines atom and coordinate generation,
explicit bond prediction, and symbolic stereochemical
reasoning. Its graph-based outputs enable atom-level
inspection and alignment with molecular images.
It reports approximately 76--93\% accuracy across
public benchmarks, demonstrating the effectiveness
of explicit structural prediction.

While these approaches provide molecular recognition
and structural supervision, PAIR additionally enforces
representation consistency across different depictions
of the same molecular structure.

\subsection{Invariant Visual Representation Learning
and Feature Adaptation}

\paragraph{Cross-View Contrastive Learning}

Cross-view contrastive learning encourages agreement between
observations sharing underlying content while preserving
discriminability. SimCLR learns representations from augmented
image views \cite{chen2020simclr}, while supervised contrastive
learning extends positive-pair supervision to samples sharing
a class label \cite{khosla2020supcon}. Wang and Isola characterize
contrastive learning through alignment and uniformity
\cite{wang2020alignment}, and theoretical work examines conditions
under which invariant content can be separated from changing
style \cite{vonkugelgen2021content}. In document recognition,
SCOB combines online text rendering with character-wise
contrastive supervision to reduce appearance differences
across domains \cite{kim2023scob}.

Related approaches have also been explored in molecular learning.
MolCLR applies contrastive learning to augmented molecular graphs
\cite{wang2022molclr}, while ImageMol combines molecular-image
contrastive learning with structural pretext tasks
\cite{zeng2022imagemol}. CGIP employs image--image and graph--image
contrastive objectives to capture chemical information
\cite{xiang2023cgip}, and AdaptMol introduces bond-level feature
alignment for domain adaptation in molecular image recognition
\cite{hu2026adaptmol}.

These studies motivate learning representations that remain
consistent across depictions of the same molecular structure.
PAIR applies this principle to paired real and rendered images,
combining molecular identity-based contrastive supervision with
recognition objectives. The aligned query representations
augment spatial image features through residual fusion,
supporting depiction consistency and chemically precise
structure reconstruction.

\paragraph{Query-Based Feature Extraction
and Residual Adaptation}

Learnable queries provide an efficient mechanism
for aggregating visual information.
BLIP-2 introduces a Q-Former that extracts compact
visual representations through cross-attention
\cite{li2023blip2}.
TokenLearner's TokenFuser reintegrates aggregated
tokens into spatial representations using residual
connections \cite{ryoo2022tokenlearner}.
Residual learning facilitates feature refinement
\cite{he2016resnet}, while CLIP-Adapter demonstrates
residual blending of adapted and pretrained
representations \cite{gao2023clipadapter}.

Inspired by these approaches, PAIR extracts
query-derived visual representations, aligns them
across molecular depictions, and integrates them
with spatial image features through residual fusion.
This combination promotes depiction consistency
while retaining the visual information required
for accurate molecular reconstruction.

\section{Methodology}

\subsection{Problem Formulation}

Suppose a chemical image is generated from molecular structure
$G$, depiction choices $A$, and noise $\eta$:
\begin{equation}
I = \operatorname{Render}(G, A) + \eta.
\end{equation}
In other words, a molecular structure $G$ defines a family
of possible images:
\begin{equation}
\mathcal{I}_G = \{I : I \text{ depicts } G\}.
\end{equation}
Images within this family share a common molecular identity:
\begin{equation}
I_1 \sim I_2 \iff G(I_1) \cong G(I_2),
\end{equation}
where $G(I)$ denotes the molecular graph depicted in image $I$,
and $\cong$ denotes graph isomorphism preserving atom and bond
attributes, including stereochemistry where specified.
Thus, equivalent images may differ in layout, drawing style,
or image quality while representing the same molecular identity.
Each family $\mathcal{I}_G$ forms an equivalence class under
this relation.

This shared identity corresponds to the same recognition target
$Y = \operatorname{CanonicalSMILES}(G)$.
Ideally, the representation preserves everything needed to
recover $G$ while ignoring irrelevant variation in $A$:
\begin{equation}
Z(I) \approx h(G).
\end{equation}
Such an ideal representation satisfies
\begin{equation}
Z(I_1) = Z(I_2)
\iff
G(I_1) \cong G(I_2).
\label{eq:structural_identity}
\end{equation}
The forward implication ensures that identical representations
correspond to the same molecular structure, while the reverse
implication ensures invariance across depictions of that structure.
Together, these conditions make the representation sufficient
to identify the molecular graph. The OCSR problem is thus viewed
as recognition over equivalence classes of drawings.

\subsection{Main Method}
\begin{figure*}[t]
    \centering
    \includegraphics[width=0.95\textwidth]{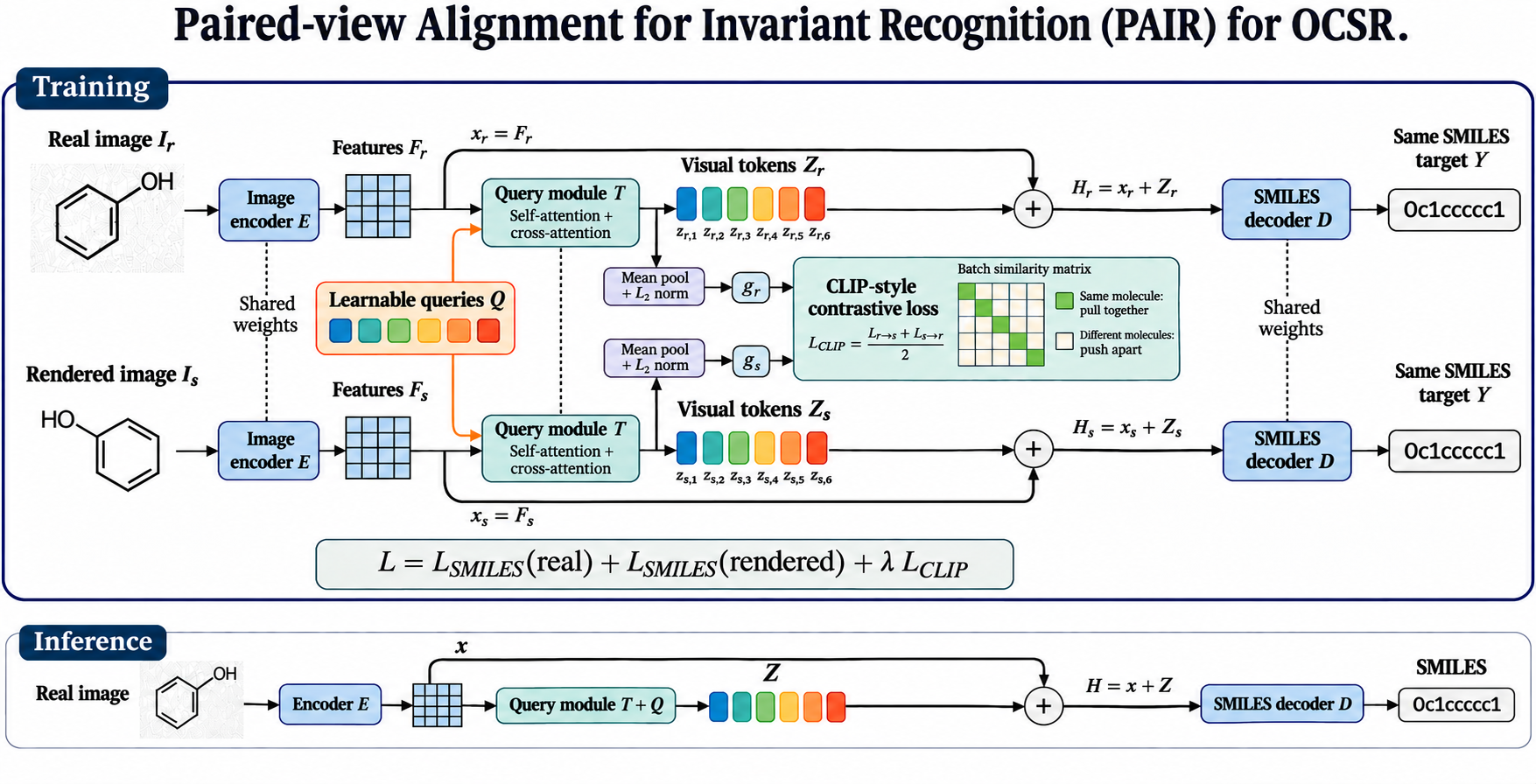}
    \caption{The framework of Depiction-Invariant Representation Learning (PAIR).}
    \label{fig:pair}
\end{figure*}

The training framework of PAIR is illusrated in \cref{fig:pair}

\subsubsection{Compute the query representations}
For each view \(v\in\{r,s\}\), the visual representations are computed as
\begin{align}
F_v &= E_\theta(I_v),\\
Z_v &= T_\phi(Q,F_v)\in\mathbb{R}^{K\times d}, \qquad Q \in \mathbb{R}^{K \times d}\\
H_v &= Z_v+F_v\in\mathbb{R}^{K\times d}.
\end{align}
in which, $E_{\theta}$ is the shared image encoder, $Q$ contains $K$ learned query embeddings, shared across views, $T_{\phi}$ uses query self-attention and cross-attention to image features, $Z_v$ contains the image-dependent query outputs, $H_v$ represents the residual-enhanced visual features, which serve as the input to the SMILES decoder.

Different from explicit chemical structural learning such as MolGrapher and MolScrbie, our work learned chemical structure implicitly; the input to the SMILES decoder are latent visual tokens, with no pre-determined atom or bond interpretation.
\subsubsection{Decode both views into SMILES}
The same autoregressive decoder is used for both views:
\begin{equation}
p_\psi(Y\mid H_v)
=
\prod_{t=1}^{L}
p_\psi\left(Y_t\mid Y_{<t},H_v\right).
\end{equation}
The SMILES reconstruction objective becomes
\begin{equation}
\mathcal{L}_{\mathrm{SMILES}}
=
-\sum_{v\in\{r,s\}}
\sum_{t=1}^{L}
\log p_\psi\left(
Y_t\mid Y_{<t},H_v
\right).
\end{equation}
Here, \(S\) denotes the ground-truth SMILES sequence and \(L\) is its token length. This decodes the molecular sequence from the latent visual representations.

\subsubsection{Alignment of $Z_r$ and $Z_s$}
To encourage consistent molecular representations across real and rendered images, we introduce a symmetric cross-view contrastive learning objective. Given the image-dependent query representations \(Z_v\) for each view \(v\in\{r,s\}\), we first aggregate the query tokens through a pooling operation, followed by a learnable projection head \(P_\omega\) and \(\ell_2\) normalization:
\begin{equation}
g_v = \operatorname{normalize}\left(P_\omega\left(\operatorname{Pool}(Z_v)\right)\right),
\end{equation}
where \(g_v\) denotes the normalized global representation of view \(v\).
Following the paired contrastive learning formulation of CLIP, we treat real and rendered images corresponding to the same molecule as positive pairs, while images associated with different molecules within a mini-batch serve as negative pairs. Specifically, for a mini-batch containing \(B\) paired samples, the real-to-rendered contrastive loss is formulated as
\begin{equation}
\mathcal{L}_{r\rightarrow s}
=
-\frac{1}{B}\sum_{i=1}^{B}
\log
\frac{
\exp\left(g_{r,i}^{\top}g_{s,i}/\tau\right)
}{
\sum_{j=1}^{B}
\exp\left(g_{r,i}^{\top}g_{s,j}/\tau\right)
},
\end{equation}
where \(\tau\) is a temperature parameter controlling the sharpness of the similarity distribution. The rendered-to-real loss \(\mathcal{L}_{s\rightarrow r}\) is defined analogously by reversing the two views. The overall symmetric contrastive objective is then given by
\begin{equation}
\mathcal{L}_{\mathrm{contrast}}
=
\frac{1}{2}
\left(
\mathcal{L}_{r\rightarrow s}
+
\mathcal{L}_{s\rightarrow r}
\right).
\end{equation}

\subsubsection{Joint Optimization of Depiction Invariance and Chemical Sufficiency}
The overall objective of our framework is to jointly learn depiction-invariant and chemically sufficient molecular representations. To this end, we combine the SMILES reconstruction objective with the cross-view contrastive alignment objective:
\begin{equation}
\begin{aligned}
\mathcal{L}_{\mathrm{total}}
=&\;
\underbrace{
\mathcal{L}_{\mathrm{SMILES}}(H_r,Y)
+
\mathcal{L}_{\mathrm{SMILES}}(H_s,Y)
}_{\text{Chemical sufficiency}}
\\
&+
\lambda
\underbrace{
\mathcal{L}_{\mathrm{contrast}}(Z_r,Z_s)
}_{\text{Depiction invariance}}.
\end{aligned}
\end{equation}
The two objectives provide complementary supervision. The weighting coefficient \(\lambda\) controls the trade-off between depiction invariance and chemical sufficiency. A larger \(\lambda\) places greater emphasis on learning consistent molecular representations across different visual depictions, whereas a smaller \(\lambda\) emphasizes preserving fine-grained chemical information necessary for accurate SMILES reconstruction. By adjusting \(\lambda\), the framework balances cross-view molecular consistency with the retention of detailed structural information.
\subsection{Experimental Design}
We evaluate PAIR from three complementary perspectives: benchmark recognition performance, representation properties, and component contributions
\subsubsection{Experimental Setup}

\paragraph{Training and model selection.}
We first adapt each backbone using a three-stage single-view
training strategy. In the first stage, only the newly introduced
Q-Former and residual gate are optimized, while the visual encoder
and language decoder remain frozen. In the second stage, the
Q-Former, residual gate, and language decoder are jointly optimized,
with the visual encoder kept frozen. In the third stage, all
components are jointly fine-tuned. Following this adaptation,
PAIR is trained on paired real and rendered molecular images,
combining recognition supervision for both views with a cross-view
contrastive objective.

Training uses MolGallery~\cite{fang2026molparserm} and the real-image supervised
fine-tuning (SFT) split of MolParser-7M~\cite{fang2025molparser}.
Within each training stage, including paired-view training,
checkpoints are selected according to the lowest validation loss
on a fixed 403-image validation split of MolParser-7M.
The checkpoint available at the beginning of each stage is
included among the candidates.

\paragraph{Evaluation datasets.}
To assess generalization across depiction sources and structural
complexity, we evaluate on six OCSR test sets covering real-world
documents, synthetic molecular depictions, and structurally
challenging molecules. The real-world benchmarks include patent
images from USPTO, chemical catalog images from UOB, and
scientific journal images from ACS~\cite{rajan2020ocsrreview,
staker2019molecular,qian2023molscribe}.
The Indigo dataset evaluates recognition on synthetically
rendered molecular structures~\cite{qian2023molscribe}.
Additionally, the Simple and Markush subsets of WildMol
assess recognition of conventional molecular structures and
more challenging Markush representations ~\cite{fang2025molparser}.
Together, these datasets provide a diverse evaluation of
recognition robustness across depiction styles, document
sources, and molecular complexity.

\paragraph{Base models.}
We instantiate PAIR on three backbones spanning different
architectures and model scales:
MolParser-Mobile-V2 (PAIR-V2; approximately 10M parameters)
~\cite{fang2025molparser},
SmolVLM-256M (PAIR-SMOL; 256M parameters)
~\cite{smolvlm},
and Qwen3-VL-2B (PAIR-2B; approximately 2B parameters)
~\cite{qwen3vl}.
MolParser-Mobile-V2 is a compact, task-specific OCSR model,
whereas SmolVLM-256M and Qwen3-VL-2B are pretrained
vision--language models.
For SmolVLM-256M, we retain its native SigLIP vision encoder,
visual connector, and SmolLM2 language model, introducing the
PAIR module without replacing these components.
This selection enables us to examine whether PAIR's benefits
extend across backbone architectures and model scales.

\subsubsection{Benchmark Recognition Performance}
\paragraph{Exact Molecular Accuracy.}
We compared PAIR with representative state-of-the-art (SOTA) OCSR methods and a matched backbone baseline. The primary evaluation metric is exact molecular accuracy:
\begin{equation}
\frac{1}{N}\sum_{i=1}^{N}
\mathbf{1}!\left[\hat G_i\cong G_i\right],
\end{equation}
where \(G_i\) and \(\hat G_i\) denote the reference and predicted SMILES expression after canonical conversion through RDKit\cite{rdkit}. Exact molecular accuracy is determined by molecular graph
equivalence after normalization and abbreviation expansion.
Stereochemistry is disregarded, while atom identities, formal
charges, isotopes, connectivity, and bond types are preserved.
For E-SMILES predictions, extended annotations must additionally
match under semantic equivalence. Invalid predictions and
unresolved comparisons are counted as incorrect.

Secondary metric is Tanimoto similarity, node recall, and edge recall. 
\paragraph{Tanimoto similarity.}
Tanimoto similarity measures fingerprint-based structural similarity between the predicted and reference molecules:
\begin{equation}
\frac{|\mathcal F(\hat G)\cap\mathcal F(G)|}
{|\mathcal F(\hat G)\cup\mathcal F(G)|},
\end{equation}
where \(\mathcal F(\cdot)\) denotes the set of active bits in a binary molecular fingerprint. 

\paragraph{Node and Edge Recall.} 
Node and edge recall quantify partial recovery of the reference molecular graph. We establish a one-to-one atom correspondence using a fixed maximum-common-substructure procedure, matching atom elements while allowing bond-order differences during alignment. Node recall is the number of matched reference atoms divided by the total number of reference atoms:
\begin{equation}
\frac{N_{\mathrm{matched\ atoms}}}{|V|}.
\end{equation}
Edge recall is the proportion of reference bonds whose mapped endpoints are connected by a predicted bond of the correct type:
\begin{equation}
\frac{N_{\mathrm{correctly\ recovered\ bonds}}}{|E|}.
\end{equation}
Thus, an incorrect bond order does not necessarily prevent atom matching, but it does count as an edge-recovery error. Because recall does not penalize additional predicted atoms or bonds, these metrics are interpreted alongside exact molecular accuracy.

\subsubsection{Representation Analysis}
\paragraph{Similarity Separation.}
We evaluate whether PAIR learns representations that remain consistent across different depictions of the same molecule while distinguishing different molecular identities. The analysis uses a fixed held-out collection of (N) real–rendered image pairs, with one pair per distinct molecular identity. All models are evaluated using the same images and retrieval candidate pool.

For molecule \(i\), we extract the query representations \(Z_{r,i},Z_{s,i}\in\mathbb{R}^{K\times d}\). Each image-level embedding is obtained by averaging the query outputs and applying L2 normalization:
\begin{align}
u_{v,i}
&= \frac{1}{K}\sum_{k=1}^{K} z_{v,i,k}
   \in \mathbb{R}^{d}, \\
g_{v,i}
&= \frac{u_{v,i}}
   {\max\left(\|u_{v,i}\|_2,\epsilon\right)}.
\end{align}
where \(\epsilon\) prevents division by zero. We then construct the cross-view cosine-similarity matrix:
\begin{equation}
S_{ij}=g_{r,i}^{\top}g_{s,j}.
\end{equation}
We report the mean matched similarity \(S^+\), mean mismatched similarity \(S^-\), and their difference $\Delta S$:
\begin{align}
S^{+}
&= \frac{1}{N}\sum_{i=1}^{N} S_{ii},\\
S^{-}
&= \frac{1}{N(N-1)}
   \sum_{i=1}^{N}\sum_{\substack{j\neq i}}^{N} S_{ij},\\
\Delta S
&= S^{+}-S^{-}.
\end{align}
Higher \(S^+\) indicates stronger cross-depiction agreement, while a larger $\Delta S$ indicates greater separation between matched and mismatched molecular pairs. These quantities are interpreted jointly: high matched similarity alone is insufficient, because embeddings of different molecules may also become similar.

\paragraph{Cross-View Retrieval.} 
We further assess whether an embedding identifies the corresponding molecular depiction among competing candidates. For real-to-rendered retrieval, each row of \(S\) is ranked in descending order. Recall at rank \(q\) is defined as:
\begin{equation}
\frac{1}{N}\sum_{i=1}^{N}
\mathbf{1}
\left[
i\in{S}[i,:q]
\right].
\end{equation}
For rendered-to-real retrieval, candidates are ranked within each column:
\begin{equation}
\frac{1}{N}\sum_{i=1}^{N}
\mathbf{1}
\left[
i\in{S}[:q,i]
\right].
\end{equation}
We report Recall@1 and Recall@5 in both directions. These metrics quantify whether molecular identity can be recovered across real and rendered appearances. 

Together, similarity separation and cross-view retrieval provide evidence of depiction-consistent molecular discrimination in the pooled query representations. They are evaluated alongside OCSR task metrics to determine whether improved representation alignment accompanies improved structural reconstruction.

\section{Results}
\label{sec:results}

\subsection{Benchmark Recognition Performance}
\label{sec:benchmark_results}

We evaluate PAIR across six OCSR benchmarks using three
model backbones. Table~\ref{tab:benchmarks} compares
PAIR variants with their corresponding baselines and
representative OCSR methods.

\paragraph{Exact molecular recognition.}
PAIR-2B achieves the highest six-dataset mean accuracy
of 85.92\%, exceeding Qwen3-VL-2B by 1.15 percentage
points. It obtains the best results on Simple (88.01\%),
Markush (62.72\%), and USPTO (95.62\%), with improvements
of 3.73, 3.12, and 0.94 percentage points over its
baseline, respectively. PAIR-V2 and PAIR-SMOL improve
mean accuracy by 0.10 and 0.21 percentage points.
Although PAIR-2B improves accuracy on five datasets,
its performance decreases on Indigo by 1.56 points,
indicating that gains vary across depiction sources.
    
\begin{table*}[t]
\centering
\caption{OCSR recognition accuracy (\%) across six
evaluation datasets without stereochemistry.
The mean is computed with equal weight across datasets.
Best results are in bold and second-best results
are underlined.}
\label{tab:benchmarks}

\begingroup
\fontsize{3.5pt}{4.5pt}\selectfont
\small
\setlength{\tabcolsep}{6pt}
\renewcommand{\arraystretch}{1.1}

\resizebox{0.85\textwidth}{!}{%
\begin{tabular}{lccccccc}
\toprule
\textbf{Model}
& \textbf{Simple}
& \textbf{Markush}
& \textbf{UOB}
& \textbf{ACS}
& \textbf{USPTO}
& \textbf{Indigo}
& \textbf{Mean} \\
& (9,997)
& (9,999)
& (5,740)
& (277)
& (5,636)
& (5,704)
& \\
\midrule

DECIMER
& 61.50 & 2.00
& \textbf{90.00} & 47.00
& 60.50 & \underline{97.50}
& 59.75 \\

SwinOCSR
& 11.50 & 1.00
& 42.50 & 26.00
& 29.50 & 79.50
& 31.67 \\

MolGrapher
& 49.62 & 1.94
& 85.21 & 58.12
& 84.60 & 95.00
& 62.42 \\

MarkushGrapher-2
& 67.78 & 6.00
& 75.50 & 66.50
& 86.50 & 74.50
& 62.80 \\

MolScribe
& 61.32 & 11.30
& 86.38 & 69.68
& \textbf{95.28} & \textbf{98.44}
& 70.40 \\

Mobile-V2
& \underline{85.91} & 53.09
& 88.97 & 86.28
& 94.04 & 96.48
& 84.13 \\

SMOL
& 78.73 & 54.44
& 88.55 & 79.42
& 87.31 & 91.95
& 80.07 \\

QWEN3-VL-2B
& 84.28 & \underline{59.60}
& 88.85 & 84.84
& 94.68 & 96.39
& \underline{84.77} \\

\midrule
\multicolumn{8}{l}{\textit{Our models}} \\

PAIR-V2
& 84.46 & 55.16
& 88.95 & \underline{86.64}
& 93.90 & 96.25
& 84.23 \\

PAIR-SMOL
& 81.85 & 56.54
& 88.78 & 79.06
& 86.09 & 89.38
& 80.28 \\

PAIR-2B
& \textbf{86.04} & \textbf{61.17}
& \underline{89.01} & \textbf{87.00}
& \underline{94.54} & 95.99
& \textbf{87.29} \\

\bottomrule
\end{tabular}%
}
\endgroup
\end{table*}

\paragraph{Partial structural recovery.}
Table~\ref{tab:structural_recovery} reports Tanimoto
similarity, node recall, and edge recall.
All PAIR variants outperform their corresponding
baselines across all three metrics.
PAIR-2B achieves the highest scores of 0.7378,
0.9401, and 0.8839, respectively.
PAIR-V2 shows particularly large improvements,
increasing Tanimoto similarity from 0.6139 to 0.7228,
node recall from 0.7609 to 0.9082, and edge recall
from 0.7038 to 0.8508.
These results demonstrate improved partial structural
recovery even when exact-match gains are modest.

\begin{table}[t]
  \centering
  \caption{Partial structural recovery measured by Tanimoto
  similarity, node recall, and edge recall.
  Higher values indicate better performance.}
  \label{tab:structural_recovery}
  \small
  \setlength{\tabcolsep}{3pt}
  \begin{tabular}{@{}lccc@{}}
    \toprule
    Model & Tanimoto $\uparrow$ & Node Recall $\uparrow$
          & Edge Recall $\uparrow$ \\
    \midrule
    Mobile-V2 & 0.6139 & 0.7609 & 0.7038 \\
    PAIR-V2   & 0.7228 & 0.9082 & 0.8508 \\
    \addlinespace
    SMOL      & 0.6790 & 0.8870 & 0.8114 \\
    PAIR-SMOL & 0.6961 & 0.9033 & 0.8373 \\
    \addlinespace
    Qwen3-VL-2B & 0.6713 & 0.8998 & 0.8080 \\
    PAIR-2B     & 0.7378 & 0.9401 & 0.8839 \\
    \bottomrule
  \end{tabular}
\end{table}

\subsection{Cross-Depiction Representation Analysis}
\label{sec:representation_results}

We assess representation consistency using matched
and mismatched similarity and bidirectional retrieval
on the same held-out molecular identities.

\paragraph{Representation discrimination.}
Table~\ref{tab:representation_results} shows that PAIR
substantially increases the similarity gap $\Delta S$
across all three backbones.
PAIR-V2, PAIR-SMOL, and PAIR-2B achieve gaps of
0.9119, 0.8797, and 0.9070, compared with 0.2936,
0.0590, and 0.3639 for their respective baselines.
The improvements are primarily driven by reduced
mismatched similarity, indicating stronger separation
between different molecular identities.

\begin{table*}[t]
  \centering
  \caption{Cross-depiction representation analysis.
  Retrieval scores are reported as percentages.}
  \label{tab:representation_results}
  \small
  \setlength{\tabcolsep}{6pt}
  \renewcommand{\arraystretch}{1.1}

  \begin{tabular}{@{}lccc*{4}{r}@{}}
    \toprule
    & & & &
    \multicolumn{2}{c}{R@1 (\%) $\uparrow$} &
    \multicolumn{2}{c}{R@5 (\%) $\uparrow$} \\
    \cmidrule(lr){5-6}
    \cmidrule(lr){7-8}
    Model & $S^{+}$ & $S^{-}$ & $\Delta S$ $\uparrow$
    & Real $\to$ Render & Render $\to$ Real
    & Real $\to$ Render & Render $\to$ Real \\
    \midrule
    Mobile-V2 & 0.9020 & 0.6083 & 0.2936
              & 85.05 & 88.40 & 95.62 & 97.16 \\
    SMOL  & 0.9952 & 0.9363 & 0.0590 
          & 97.16 & 96.13 & 99.48 & 98.97 \\
    Qwen3-VL  & 0.9136 & 0.5497 & 0.3639 & 87.89 & 82.47 & 98.71 & 92.78 \\
    \midrule
    PAIR-V2   & 0.9272 & 0.0153 & 0.9119
              & 98.71 & 99.23 & 100.00 & 99.74 \\
    PAIR-SMOL & 0.9345 & 0.0548 & 0.8797 
              & 98.97 & 99.23 & 99.74 & 99.74 \\
    PAIR-2B  & 0.9314 & 0.0243 & 0.9070 
             & 97.16 & 97.68 & 99.74 & 98.45 \\
    \bottomrule
  \end{tabular}
\end{table*}

\begin{figure}[t]
    \centering
    \includegraphics[width=0.5\textwidth]{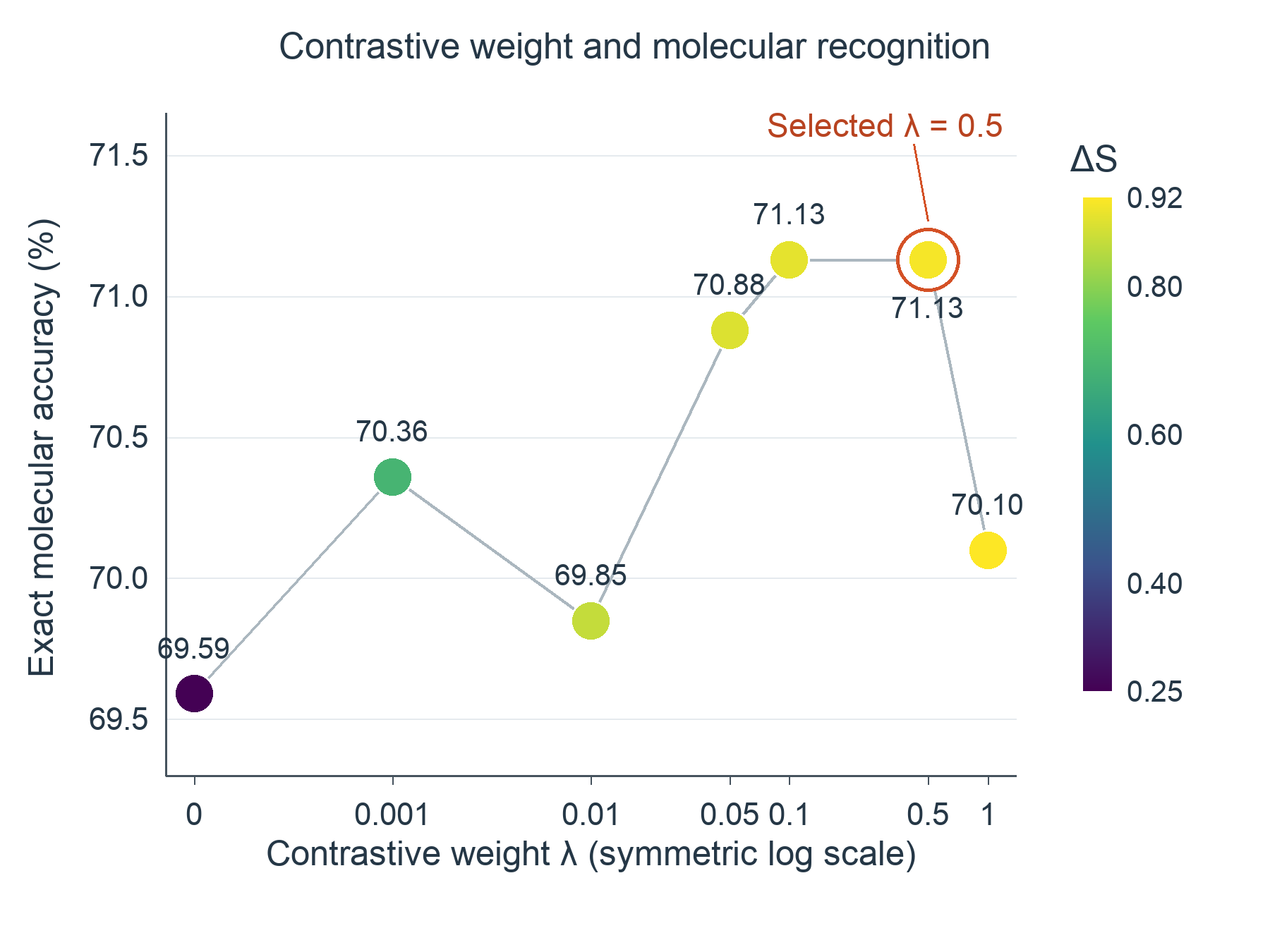}
    \caption{The framework of Depiction-Invariant Representation Learning (PAIR).}
    \label{fig:lambda}
\end{figure}

Furthermore, Figure~\ref{fig:lambda} reports recognition accuracy
and representation metrics for
$\lambda\in\{0,0.001,0.01,0.05,0.1,0.5,1\}$.
The best validation-selected configuration uses
$\lambda=0.5$. While the depiction-alignment strength clearly increases with higher $\lambda$, a hyperbolic relationship is identified between the depiction-invariance and chemicacal sufficiency, indicating the trade-off exits empirically. This distinction is important because an increase in matched
similarity alone does not establish molecular discrimination.

\paragraph{Bidirectional cross-view retrieval.}
PAIR improves Recall@1 and Recall@5 in both retrieval
directions across all backbones.
PAIR-V2 achieves 98.71\% real-to-rendered Recall@1
and 99.23\% rendered-to-real Recall@1, compared with
85.05\% and 88.40\% for Mobile-V2.
PAIR-SMOL achieves 98.97\% and 99.23\%, while
PAIR-2B achieves 97.16\% and 97.68\%, respectively.
These results confirm improved correspondence between
real and rendered molecular representations.

\subsection{Ablation Studies}
\label{sec:ablation_results}

We conduct controlled ablations to isolate the contributions
of paired-image supervision, residual query refinement,
and cross-view contrastive alignment. 
Table~\ref{tab:component_ablation} compares:
(B0) the backbone trained on real images;
(B1) the backbone trained on real and rendered images;
(B2) residual query refinement without contrastive alignment;
and (B3) full PAIR.

\begin{table}[t]
  \centering
  \caption{Component ablation of PAIR.
  Paired denotes recognition supervision on both real and rendered
  images; Residual denotes query-based residual refinement;
  Contrastive denotes cross-view contrastive alignment.
  Exact molecular accuracy (EMA) is reported in percentages.
  All configurations use the same base model Mobile-V2.}
  \label{tab:component_ablation}
  \small
  \setlength{\tabcolsep}{4pt}
  \begin{tabular}{@{}lcccc@{}}
    \toprule
    Variant
    & Paired
    & Residual
    & Contrastive
    & \makecell{EMA (\%)\\$\uparrow$} \\
    \midrule
    B0 & --         & --         & --         & 62.63 \\
    B1 & $\checkmark$ & --         & --         & 65.21 \\
    B2 & $\checkmark$ & $\checkmark$ & --         & 69.59 \\
    B3 (PAIR)
       & $\checkmark$ & $\checkmark$ & $\checkmark$ & 71.13 \\
    \bottomrule
  \end{tabular}
\end{table}

\paragraph{Component contributions.}
Paired-image supervision increases exact molecular
accuracy from 62.63\% to 65.21\%, a gain of
2.58 percentage points. Adding residual query
refinement further improves accuracy to 69.59\%
(+4.38 points). Incorporating cross-view contrastive
alignment raises accuracy to 71.13\%
(+1.54 points). Overall, full PAIR improves accuracy
by 8.50 percentage points over the real-image-only
baseline. These results demonstrate complementary
benefits from all three components, with residual
refinement providing the largest incremental gain
in this ablation.

\section{Discussion}
\label{sec:discussion}

The experiments demonstrate that PAIR improves molecular
structure recognition across different backbone architectures.
PAIR-2B achieves the highest mean exact-match accuracy across
six benchmarks, while all three PAIR variants consistently
improve partial structural recovery and cross-depiction retrieval.
The ablation studies further confirm the complementary
contributions of paired-image supervision, residual query
refinement, and contrastive alignment. Although exact-match
improvements vary across datasets, these results support
the effectiveness of cross-depiction representation learning
for OCSR.

\paragraph{Invariance Requires Structural Discrimination.}
The findings provide empirical support for the theoretical
motivation introduced in Section~\ref{sec:intro}: an ideal
molecular representation should remain invariant across
different depictions of the same graph while distinguishing
different molecular graphs. These two conditions correspond
to within-molecule consistency and between-molecule
discrimination. PAIR approximates them through cross-view
contrastive learning, substantially increasing the
matched--mismatched similarity gap, primarily by reducing
similarity between distinct molecules. However, invariance
alone is insufficient for accurate molecular reconstruction
if chemically informative visual distinctions are lost.
The contrastive-weight analysis reveals this trade-off:
stronger alignment does not necessarily improve recognition
accuracy. Residual query refinement addresses this tension
by retaining spatial image features alongside aligned query
representations, while SMILES supervision encourages the
preservation of molecular information. Together, these
mechanisms balance depiction invariance with chemical
sufficiency, although the experiments do not establish
the ideal theoretical conditions formally.

\paragraph{Consistent Gains in Markush Recognition.}
PAIR demonstrates particularly consistent improvements
on Markush structures, with gains across all three
backbones and PAIR-2B achieving 62.72\% accuracy.
Markush representations introduce additional complexity
through variable substituents, generic groups, and
nonstandard structural annotations, increasing the
diversity of visual expressions. We hypothesize that
cross-depiction alignment helps reduce sensitivity to
irrelevant drawing variations, while residual refinement
preserves the local structural details needed to
distinguish molecular identities. This combination may
be especially beneficial for visually heterogeneous
Markush representations. Although PAIR does not
explicitly model Markush semantics, the consistent
improvements suggest that depiction-invariant learning
offers a promising approach to this challenging
recognition setting.

\paragraph{Limitations and future directions.}
PAIR does not guarantee complete depiction invariance
or unique representations for all molecular graphs,
and its effectiveness remains dependent on backbone
architecture, dataset characteristics, and contrastive
strength. Moreover, improvements in partial structural
recovery do not always translate into exact molecular
predictions. The current evaluation also excludes
stereochemistry, leaving its preservation under
cross-depiction alignment unexplored. Future work
could investigate chemically informed contrastive
objectives that better distinguish structurally similar
molecules, adaptive alignment strategies that preserve
fine-grained chemical information, and extensions to
stereochemistry and more complex molecular
representations.

{
    \small
    \bibliographystyle{ieeenat_fullname}
    \bibliography{main}
}

\end{document}